# Brain decoding from functional MRI using long short-term memory recurrent neural networks

Hongming Li, Yong Fan

Center for Biomedical Image Computing and Analytics,
Department of Radiology, Perelman School of Medicine, University of Pennsylvania

**Abstract.** Decoding brain functional states underlying different cognitive processes using multivariate pattern recognition techniques has attracted increasing interests in brain imaging studies. Promising performance has been achieved using brain functional connectivity or brain activation signatures for a variety of brain decoding tasks. However, most of existing studies have built decoding models upon features extracted from imaging data at individual time points or temporal windows with a fixed interval, which might not be optimal across different cognitive processes due to varying temporal durations and dependency of different cognitive processes. In this study, we develop a deep learning based framework for brain decoding by leveraging recent advances in sequence modeling using long short-term memory (LSTM) recurrent neural networks (RNNs). Particularly, functional profiles extracted from task functional imaging data based on their corresponding subject-specific intrinsic functional networks are used as features to build brain decoding models, and LSTM RNNs are adopted to learn decoding mappings between functional profiles and brain states. We evaluate the proposed method using task fMRI data from the HCP dataset, and experimental results have demonstrated that the proposed method could effectively distinguish brain states under different task events and obtain higher accuracy than conventional decoding models.

## 1    Introduction

Decoding the brain based on functional signatures derived from imaging data using multivariate pattern recognition techniques has become increasingly popular in recent years. With the massive spatiotemporal information provided by the functional brain imaging data, such as functional magnetic resonance imaging (fMRI), several strategies have been proposed for the brain decoding [1-7].

Most of the existing fMRI based brain decoding studies focus on identification of functional signatures that are informative for distinguishing different brain states. Particularly, brain activations evoked by task stimuli identified using a general linear model (GLM) framework are commonly adopted [8]. The procedure of identifying brain activation maps is equivalent to a supervised feature selection procedure, which may improve the sensitivity of the brain decoding. In addition to feature selection using the GLM framework, several studies select regions of interests (ROIs) related to the brain decoding tasks based on *a prior* anatomical/functional knowledge [2]. A two-step strategy [4] that swaps the functional signature identification from spatial domain to temporal domain has recently been proposed to decode fMRI activity in the time domain, aiming to overcome the curse of dimensionality problem caused by spatial functional signatures used for the brain decoding. All these aforementioned methods require knowledge of timing information of task events or types of tasks to carry out the feature selection for the brain decoding, which limits their general application. Other than task-specific functional signatures identified in a supervised manner, several whole-brain functional signatures have been proposed. In particular, whole-brain functional connectivity patterns based on resting-state brain networks identified using independent component analysis (ICA) are adopted for the brain decoding [1]. However, time windows with a properly defined width are required in order to reliably estimate the functional connectivity patterns. Deep belief neural network (DBN) has been adopted to learn a low-dimension representation of 3D fMRI volume for the brain decoding [3], where 3D images are flatten into 1D vectors as features for learning the DBN, losing the spatial structure information of the 3D images. More recently, 3D convolutional neural networks (CNNs) are adopted to learn a latent representation for decoding functional brain task states [5]. Although the CNNs could learn discriminative representations effectively, it is nontrivial to interpret biological meanings of the learned features.

Most of the existing studies perform the brain decoding based on functional signatures computed at individual time points or temporal windows with a fixed length using conventional classification techniques, such as support vector machine (SVM) [9] and logistic regression [2, 4]. These classifiers do not take into consideration the temporal dependency, which is inherently available in the sequential fMRI data and may boost the brain decoding performance. Though functional signatures extracted from time windows [1, 4, 5] may help capture the temporal dependency implicitly, time windows with a fixed width are not necessarily optimal over different brain states since they may change at unpredictable intervals. On the other hand, recurrent neural networks (RNNs) with long short-term memory (LSTM) [10] have achieved remarkable advances in sequence modeling [11], and these techniques might be powerful alternatives for the brain decoding tasks.

In this study, we develop a deep learning based framework for decoding the brain states from task fMRI data, by leveraging recent advances in RNNs. Particularly, we learn mappings between functional signatures and brain states by adopting LSTM RNNs which could capture the temporal dependency adaptively by learning from data. Instead of selecting ROIs or fMRI features using feature selection techniques or *a prior* knowledge of problems under study, we extract functional profiles from task functional imaging data based on subject-specific intrinsic functional networks and the functional profiles are used as features for building LSTM RNNs based brain decoding models. Our method has been evaluated for predicting brain states based on task fMRI data obtained from the human connectome project (HCP) [12], and experimental results have demonstrated that the proposed method could obtain better brain decoding performance than the conventional methods.

## 2 Methods

To decode the brain state from task fMRI data, a prediction model of LSTM RNNs [10] is trained based on functional signatures extracted using a functional brain decomposition technique [13, 14]. The overall framework is illustrated in Fig. 1(a).

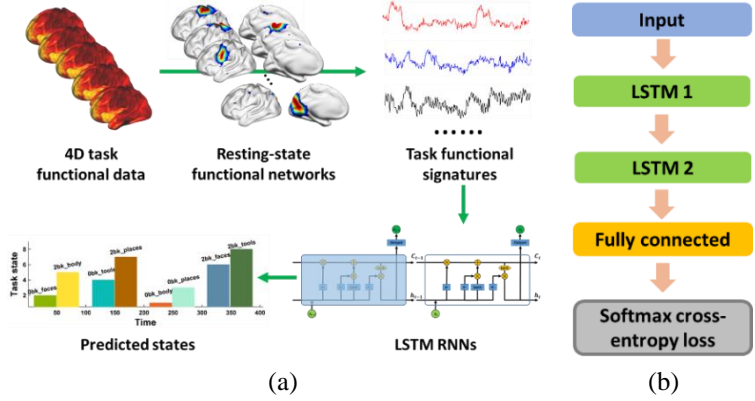

(a)                           (b)

**Fig. 1.** Schematic illustration of the proposed brain decoding framework. (a) The overall architecture of the proposed model, (b) LSTM RNNs used in this study.

### 2.1 Functional signature based on intrinsic functional networks

With good correspondence to the task activations [15], intrinsic functional networks (FNs) provided an intuitive and generally applicable means to extract functional signatures for the brain state decoding. Using the FNs, 3D fMRI data could be represented by a low-dimension feature vector, which could alleviate the curse of dimensionality, be general to different brain decoding tasks, and provide better interpretability. Instead of identifying ROIs at a group level [1], we applied a collaborative sparse brain decomposition model [13, 14] to the resting-state fMRI data of all the subjects used for the brain decoding to identify subject-specific FNs.

Given a group of $n$ subjects, each having a resting-state fMRI scan $D^i \in R^{T \times S}$, $i = 1, 2, \ldots, n$, consisting of $S$ voxels and $T$ time points, we first obtain $K$ FNs $V^i \in R_+^{K \times S}$ and its corresponding functional time courses $U^i \in R^{T \times K}$ for each subject using the collaborative sparse brain decomposition model [13, 14], which could identify subject-specific functional networks with inter-subject correspondence and better characterize the intrinsic functional representation at an individual subject level. Based on the subject-specific FNs, the functional signatures $F^i \in R^{T \times K}$ used for the brain decoding are defined as weighted mean time courses of the task fMRI data within individual FNs, and are calculated by

$$F^i = D_f^i \cdot \left(V_N^i\right)', \qquad (1)$$

where $D_f^i$ is the task fMRI data of subject $i$ for the brain decoding, $V_N^i$ is the row-wise normalized $V^i$ with its row-wise sum equal to one. Example FNs used in our study are illustrated in Fig. 2.

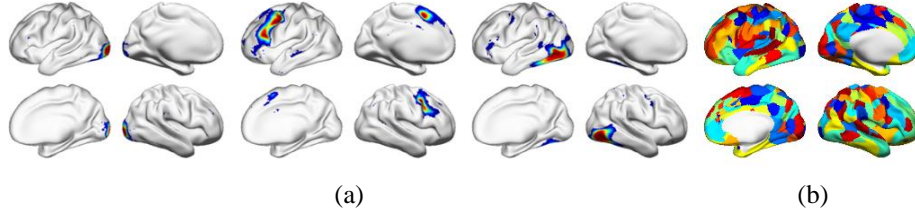

**Fig. 2.** Functional networks used to extract task functional signatures for the brain decoding. (a) Example functional networks, (b) all functional networks encoded in different colors.

### 2.2 Brain decoding using LSTM RNNs

Given the functional signatures $F^i$ of a group of $n$ subjects, $i = 1,2,...,n$, a LSTM RNNs [10] model is built to predict the brain state of each time point based on its functional profile and temporal dependency on its preceding time points. The architecture of the LSTM RNNs used in this study is illustrated in Fig. 1(b), including two hidden LSTM layers and one fully connected layer. Two hidden LSTM layers are used to encode the functional information with temporal dependency for each time point, and the fully connected layer is used to learn a mapping between the learned feature representation and the brain states. The functional representation encoded in each LSTM layer is calculated as

$$\begin{aligned}
f_t^l &= \sigma(W_f^l \cdot [h_{t-1}^l, x_t^l] + b_f^l), \\
i_t^l &= \sigma(W_i^l \cdot [h_{t-1}^l, x_t^l] + b_i^l), \\
\tilde{C}_t^l &= tanh(W_C^l \cdot [h_{t-1}^l, x_t^l] + b_c^l), \\
C_t^l &= f_t^l * C_{t-1}^l + i_t^l * \tilde{C}_t^l, \\
o_t^l &= \sigma(W_o^l \cdot [h_{t-1}^l, x_t^l] + b_o^l), \\
h_t^l &= o_t^l * tanh(C_t^l),
\end{aligned} \quad (2)$$

where $f_t^l$, $i_t^l$, $C_t^l$, $h_t^l$, and $x_t^l$ denote the output of forget gate, input gate, cell state, hidden state, and the input feature vector of the $l$-th LSTM layer ($l = 1, 2$) at the $t$-th time point respectively, and $\sigma$ denotes the sigmoid function. The input features to the first LSTM layer are the functional signatures derived from FNs, and the input to the second LSTM layer is a hidden state vector obtained by the first LSTM layer. A fully connected layer with $S$ output nodes is adopted for predicting the brain state as

$$s_t = softmax(W_s \cdot h_t^2 + b_s), \quad (3)$$

where $S$ is the number of brain states to be decoded, and $h_t^2$ is the hidden state output of the second LSTM layer which encodes the input functional signature at the $t$-th time point and the temporal dependency information encoded in the cell state from its preceding time points.

In this study, each hidden LSTM layer contains 256 hidden nodes, and softmax cross-entropy between real and predicted brain states is used as the objective function to optimize the LSTM RNNs model.

## 3 Experimental results

We evaluated the proposed method based on task and resting-state fMRI data of 490 subjects from the HCP [12]. In this study, we focused on the working memory task, which consisted of 2-back and 0-back task blocks of tools, places, faces and body, and a fixation period. Each working memory fMRI scan consisted of 405 time points of 3D volumes, and its corresponding resting-state fMRI scan had 1200 time points. The fMRI data acquisition and task paradigm were detailed in [12].

We applied the collaborative sparse brain decomposition model [13, 14] to the resting-state fMRI data of 490 subjects for identifying 90 subject-specific FNs. The number of FNs was estimated by MELODIC [16]. The subject-specific FNs were then used to extract functional signatures of the working memory task fMRI data for each subject, which was a matrix of 405 by 90. The proposed method was then applied to the functional signatures to predict their corresponding brain states. Particularly, we split the whole dataset into training, validation, and testing datasets. The training dataset included data of 400 subjects for training the LSTM RNNs model, the validation dataset included data of 50 subjects for determining the early-stop of the training procedure, and data of the remaining 40 subjects were used as an external testing dataset.

Due to the delay of blood oxygen level dependent (BOLD) response observed in fMRI data, the occurrence of brain response is typically not synchronized with the presentation of stimuli, so the brain state for each time point was adjusted according to the task paradigm and the delay of BOLD signal before training the brain decoding models. Based on an estimated BOLD response delay of 6s [17], we shifted the task paradigms forward by 8 time points and used them to update the ground truth brain states for training and evaluating the proposed brain state decoding model.

To train a LSTM RNNs model, we have generated training samples by cropping the functional signatures of each subject into clip matrices of 40 by 90, with an overlap of 20 time points between temporally consecutive training clips. We adopted the cropped dataset for training our model for following reasons. Firstly, the task paradigms of most subjects from the HCP dataset shared almost the identical temporal patterns. In other words, the ground truth brain states of most subjects were the same, which may mislead the model training to generate the same output regardless of the functional signatures fed into the LSTM RNNs model if we used their full-length data for training the brain decoding model. In our study, the length of data clips was set to 40 so that each clip contained 2 or 3 different brain states and such randomness could eliminate the aforementioned bias. Secondly, the data clips with temporal overlap also served as data augmentation of the training samples for improving the model training. When evaluating our LSTM RNNs model, we applied the trained model to the full-length functional signatures of the testing subjects to predict brain states of their entire task fMRI scans. We implemented the proposed method using Tensorflow. Particularly, we adopted the ADAM optimizer with a learning rate of 0.001, which was updated every 50,000 training steps with a decay rate of 0.1, and the total number of training steps was set to 200,000. Batch size was set to 32 during the training procedure.

We compared the proposed model with a brain decoding model built using random forests [18], which used the functional signatures at individual time points as features. The random forests classifier was adopted due to its inherent feature selection mechanism and its capability of handling multi-class classification problems. For the random forests based brain decoding model, the number of decision trees and the minimum leaf size of the tree were selected from a set of parameters ({100, 200, 500, 1000} for the number of trees, and {3, 5, 10} for the minimum leaf size) to optimize its brain decoding performance based on the validation dataset.

### 3.1 Brain decoding on working memory task fMRI data

The mean normalized confusion matrices of the brain decoding accuracy on the 40 testing subjects obtained by the random forests and the LSTM RNNs models are shown in Fig. 3. The LSTM RNNs model outperformed the random forests model in 5 out of 9 brain states (Wilcoxon signed rank test, $p < 0.002$). The overall accuracy obtained by the LSTM RNNs model was $0.687 \pm 0.371$, while the overall accuracy obtained by the random forests model was $0.628 \pm 0.234$, demonstrating that our method performed significantly better than the random forests based prediction models (Wilcoxon signed rank test, $p < 0.001$). The improved performance indicates that the temporal dependency encoded in the LSTM RNNs model could provide more discriminative information for the brain decoding.

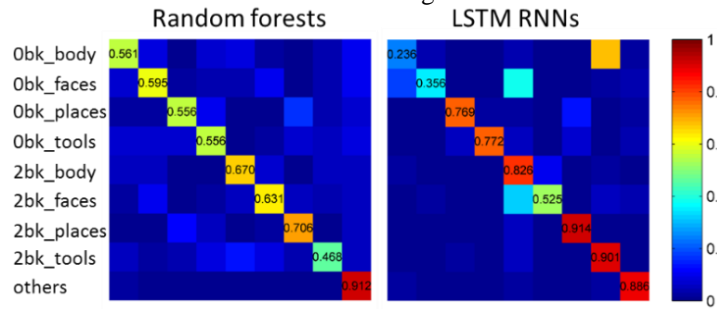

**Fig. 3.** Brain decoding performance of the random forests and LSTM RNNs models on the testing dataset of working memory task fMRI. The colorbar indicates mean decoding accuracy on the 40 testing subjects.

### 3.2 Sensitivity analysis of the brain decoding model

To understand the LSTM RNNs based decoding model, we have carried out a sensitivity analysis to determine how changes in the functional signatures affect the decoding model based on the 40 testing subjects using a principal component analysis (PCA) based sensitivity analysis method [19]. Particularly, with the trained LSTM RNNs model fixed, functional signatures of 90 FNs were excluded (i.e., their values were set to zero) one by one from the input and changes in the decoding accuracy were recorded. Once all the changes in the brain decoding accuracy with respect to all FNs were obtained for all testing subjects, we obtained a change matrix of $90 \times 40$, encapsulating changes of the brain decoding. We then applied PCA to the change matrix to identify principle components (PCs) that encoded main directions of the prediction changes with respect to changes in the functional signatures of FNs.

The sensitive analysis revealed FNs whose functional signatures were more sensitive than others to the brain decoding on the working memory task fMRI data. Particularly, among top 5 FNs with the largest magnitudes in the first PC as shown in Fig. 4, four of them were corresponding to the working memory evoked activations as demonstrated in [20], indicating that the LSTM RNNs model captured the functional dynamics of the working memory related brain states.

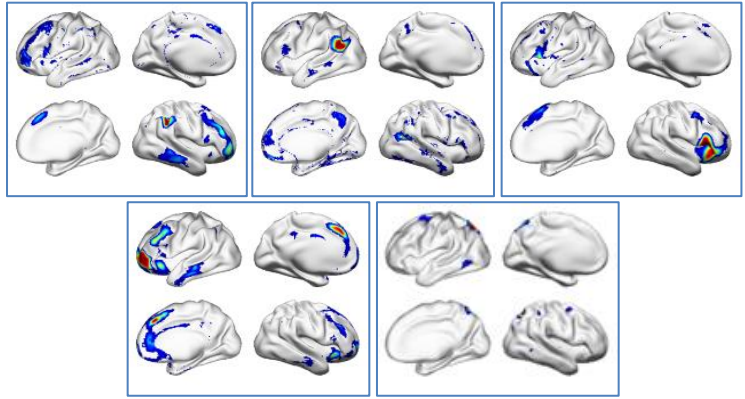

**Fig. 4.** Sensitivity analysis of the brain decoding model on the working memory task fMRI dataset. The top 5 FNs with most sensitive functional signatures are illustrated.

## 4   Conclusions

In this study, we propose a deep learning based model for decoding the brain states underlying different cognitive processes from task fMRI data. Subject-specific intrinsic functional networks are used to extract task related functional signatures, and the LSTM RNNs technique is adopted to adaptively capture the temporal dependency within the functional data as well as the relationship between the learned functional representations and the brain functional states. The experimental results on the working memory task fMRI dataset have demonstrated that the proposed model could obtain improved brain decoding performance compared with a decoding model without considering the temporal dependency.

## 5   Acknowledgements

This work was supported in part by National Institutes of Health grants [CA223358, EB022573, DK114786, DA039215, and DA039002] and a NVIDIA Academic GPU grant.

## References

1. Shirer, W.R., et al., *Decoding subject-driven cognitive states with whole-brain connectivity patterns.* Cereb Cortex, 2012. **22**(1): p. 158-65.

2.  Huth, A.G., et al., *Decoding the Semantic Content of Natural Movies from Human Brain Activity.* Front Syst Neurosci, 2016. **10**: p. 81.
3.  Jang, H., et al., *Task-specific feature extraction and classification of fMRI volumes using a deep neural network initialized with a deep belief network: Evaluation using sensorimotor tasks.* Neuroimage, 2017. **145**(Pt B): p. 314-328.
4.  Loula, J., G. Varoquaux, and B. Thirion, *Decoding fMRI activity in the time domain improves classification performance.* Neuroimage, 2017.
5.  Wang, X., et al., *Task state decoding and mapping of individual four-dimensional fMRI time series using deep neural network.* arXiv preprint arXiv:1801.09858, 2018.
6.  Fan, Y., D. Shen, and C. Davatzikos. *Detecting Cognitive States from fMRI Images by Machine Learning and Multivariate Classification*. in *2006 Conference on Computer Vision and Pattern Recognition Workshop (CVPRW'06)*. 2006.
7.  Davatzikos, C., et al., *Classifying spatial patterns of brain activity with machine learning methods: Application to lie detection.* Neuroimage, 2005. **28**(3): p. 663-668.
8.  Mumford, J.A., et al., *Deconvolving BOLD activation in event-related designs for multivoxel pattern classification analyses.* Neuroimage, 2012. **59**(3): p. 2636-43.
9.  Shen, G., et al., *Decoding the individual finger movements from single-trial functional magnetic resonance imaging recordings of human brain activity.* Eur J Neurosci, 2014. **39**(12): p. 2071-82.
10. Hochreiter, S. and J. Schmidhuber, *Long short-term memory.* Neural computation, 1997. **9**(8): p. 1735-1780.
11. Lipton, Z.C., J. Berkowitz, and C. Elkan, *A critical review of recurrent neural networks for sequence learning.* arXiv preprint arXiv:1506.00019, 2015.
12. Glasser, M.F., et al., *The minimal preprocessing pipelines for the Human Connectome Project.* Neuroimage, 2013. **80**: p. 105-24.
13. Li, H., T.D. Satterthwaite, and Y. Fan, *Large-scale sparse functional networks from resting state fMRI.* Neuroimage, 2017. **156**: p. 1-13.
14. Li, H., T. Satterthwaite, and Y. Fan. *Identification of subject-specific brain functional networks using a collaborative sparse nonnegative matrix decomposition method*. in *2016 IEEE 13th International Symposium on Biomedical Imaging (ISBI)*. 2016.
15. Smith, S.M., et al., *Correspondence of the brain's functional architecture during activation and rest.* Proc Natl Acad Sci U S A, 2009. **106**(31): p. 13040-5.
16. Jenkinson, M., et al., *Fsl.* Neuroimage, 2012. **62**(2): p. 782-90.
17. Liao, C.H., et al., *Estimating the delay of the fMRI response.* Neuroimage, 2002. **16**(3 Pt 1): p. 593-606.
18. Breiman, L., *Random forests.* Machine learning, 2001. **45**(1): p. 5-32.
19. Koyamada, S., et al. *Principal Sensitivity Analysis*. in *Pacific-Asia Conference on Knowledge Discovery and Data Mining*. 2015. Springer.
20. Barch, D.M., et al., *Function in the human connectome: task-fMRI and individual differences in behavior.* Neuroimage, 2013. **80**: p. 169-89.